\documentclass{llncs}
\usepackage{makeidx}  
\usepackage{graphicx}
\usepackage{subfigure}
\usepackage{amssymb,amsmath,bm}
\usepackage{mathrsfs}
\usepackage{booktabs}
\usepackage{multirow}
\begin{document}
	\frontmatter          
	\title{Combating Uncertainty with Novel Losses for Automatic Left Atrium Segmentation}
	
	\author{Xin Yang$^{1,4\dagger*}$ \and Na Wang$^{2,3\dagger}$ \and Yi Wang$^{2,3}$ \and Xu Wang$^{2,3}$ \and Reza Nezafat$^{4}$ \and \\Dong Ni$^{2,3}$ \and Pheng-Ann Heng$^{1}$}
	
	\institute{Dept. of Computer Science and Engineering, The Chinese University of Hong Kong\\
		\textit{xinyang@cse.cuhk.edu.hk}\and
		National-Regional Key Technology Engineering Laboratory for Medical Ultrasound, \\School of Biomedical Engineering, Health Science Center, Shenzhen University, China\and
		Medical UltraSound Image Computing (MUSIC) Lab\and
		Department of Medicine (Cardiovascular Division), Beth Israel Deaconess Medical Center and Harvard Medical School}
	
	\maketitle
	\let\thefootnote\relax\footnotetext{$\dagger$Authors contribute equally to this work. *The work was partly finished when Xin Yang was an internship in the Department of Medicine (Cardiovascular Division), Beth Israel Deaconess Medical Center and Harvard Medical School.}
	\addtocounter{footnote}{-1}\let\thefootnote\svthefootnote
	
	\begin{abstract}
		 Segmenting left atrium in MR volume holds great potentials in promoting the treatment of atrial fibrillation. However, the varying anatomies, artifacts and low contrasts among tissues hinder the advance of both manual and automated solutions. In this paper, we propose a fully-automated framework to segment left atrium in gadolinium-enhanced MR volumes. The region of left atrium is firstly automatically localized by a detection module. Our framework then originates with a customized 3D deep neural network to fully explore the spatial dependency in the region for segmentation. To alleviate the risk of low training efficiency and potential overfitting, we enhance our deep network with the transfer learning and deep supervision strategy. Main contribution of our network design lies in the composite loss function to combat the boundary ambiguity and hard examples. We firstly adopt the \textit{Overlap} loss to encourage network reduce the overlap between the foreground and background and thus sharpen the predictions on boundary. We then propose a novel \textit{Focal Positive} loss to guide the learning of voxel-specific threshold and emphasize the foreground to improve classification sensitivity. Further improvement is obtained with an recursive training scheme. With ablation studies, all the introduced modules prove to be effective. The proposed framework achieves an average Dice of $92.24\%$ in segmenting left atrium with pulmonary veins on 20 testing volumes. \par
	\end{abstract}

	\section{Introduction}
	Cardiovascular diseases are keeping as the leading causes of death in the world. About half of the diagnosed cases suffer from the atrial fibrillation caused stroke \cite{peng2016review}. MR, especially the Gadolinium Enhancement MR (GE-MR), is a dominant imaging modality to localize scars and provide guidance for ablation therapy \cite{mcgann2013atrial}. Segmenting the left atrium with associated pulmonary veins and reconstructing its patient-specific anatomy are beneficial to optimize the therapy plan and reduce the risk of intervention. However, manual delineation tends to be time-consuming and presents low inter- and intra-expert reproducibilities. \par
	
\begin{figure}[h]
		\centering
		\vspace{-0.5mm}
		\includegraphics[width=0.88\linewidth]{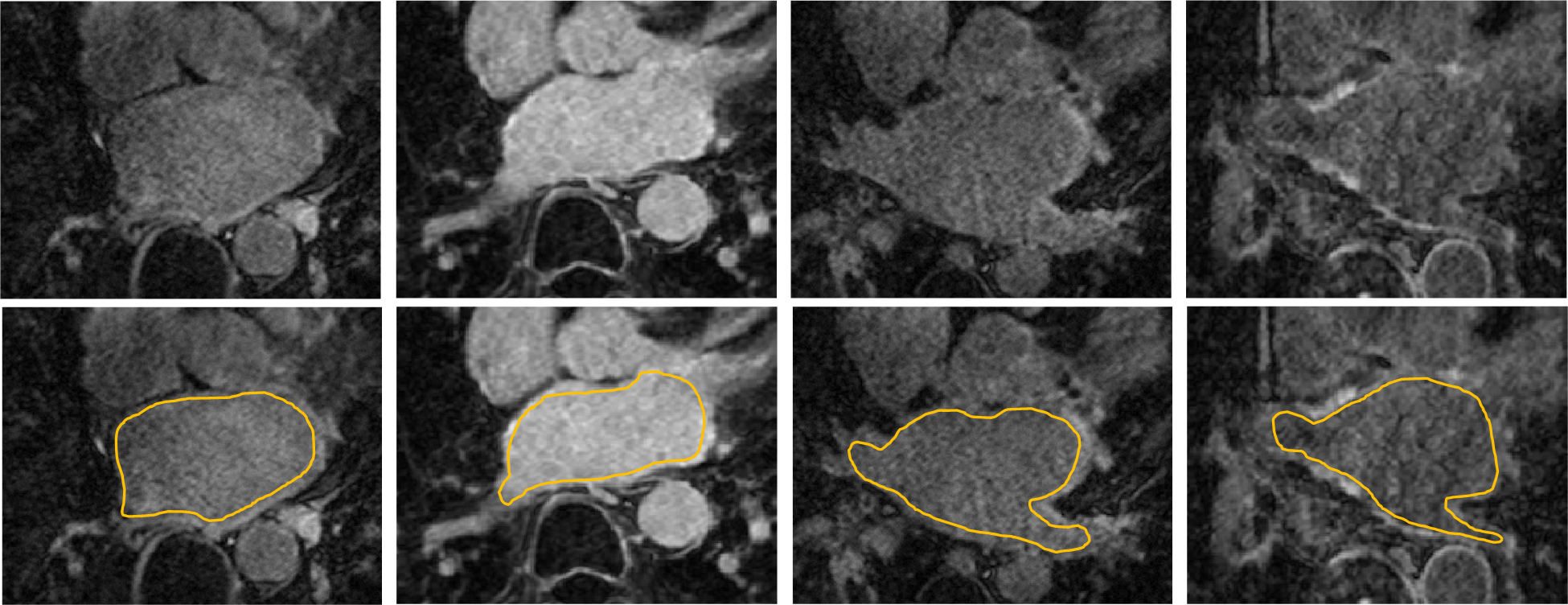}
		\caption{Challenges in segmenting the left atrium and associated pulmonary veins. Yellow curve denotes segmentation ground truth.}
		\label{fig:challenge}
		\vspace{-0.6mm}
	\end{figure}

	Automatically segmenting left atrium with pulmonary veins is a nontrivial task. As shown in Fig. \ref{fig:challenge}, left atrium and pulmonary veins vary greatly in shape and size across subjects. Affected by the artifacts and differences in gadolinium dose, appearance of atrium and pulmonary veins can change dramatically. Due to the thin myocardial wall of left atrium and the low contrast between left atrium and other surroundings, the boundary of left atrium and pulmonary veins are hard and ambiguous for computer to recognize. Conquering the complex anatomical variance and boundary ambiguity are the main concerns of this work. \par
	
	Cardiovascular image segmentation is an active research area. Deformable models were proposed to segment chambers and vessels \cite{zheng2014multi}. Other popular streams are the multi-atlas \cite{bai2015multi} and non-rigid registration based methods \cite{zhuang2010registration}. However, those methods bear the difficulties in designing boundary descriptors and generalizing the model trained on limited data to unseen deformable cases \cite{tobon2015benchmark}. Deep neural network rapidly emerges and pushes the upper-bound of cardiovascular image segmentation \cite{chen2018multiview,mortazi2017cardiacnet,yang2017fully}. Whereas, limited by computational resources, most methods exploit 2D architectures which ignore the global spatial information in volume and the segmentation consistency across slices. Also, few works paid attention to the importance of loss function design. Hybrid loss function combining \textit{Dice} loss and weighted cross entropy loss can address class-imbalance and preserve both concrete and branchy structures in segmentation \cite{yang2017hybrid}. Loss function enforcing the low overlap between foreground and background, denoted as \textit{Overlap} loss, presents potential in promoting segmentation \cite{wang2018automated}. \par
	
	In this paper, we propose a fully-automated framework to segment left atrium and pulmonary veins in gadolinium-enhanced MR volumes. We firstly deploy a detection module to accurately localize the left atrium region in the raw volume. We then propose our customized segmentation network in 3D fashion. Transfer learning and dense deep supervision strategy are involved to alleviate the risk of low training efficiency and potential overfitting. To effectively guide our network combat the boundary ambiguity and hard examples, we introduce a composite loss function which consists of two complementary components: 1) the \textit{Overlap} loss to encourage network reduce the intersection between foreground and background probability maps and thus make the predictions on boundary less fuzzy, 2) a novel \textit{Focal Positive} loss to guide the learning of voxel-specific threshold and emphasize the foreground to improve classification sensitivity. Based on the design, we obtain further improvement by fine-tuning our network with a recursive scheme. With ablation studies, all the introduced modules prove to be effective. The proposed framework achieves an average Dice of $92.24\%$ in segmenting left atrium with pulmonary veins on 20 testing volumes.	
	\begin{figure}[h]
		\centering
		\includegraphics[width=1.0\linewidth]{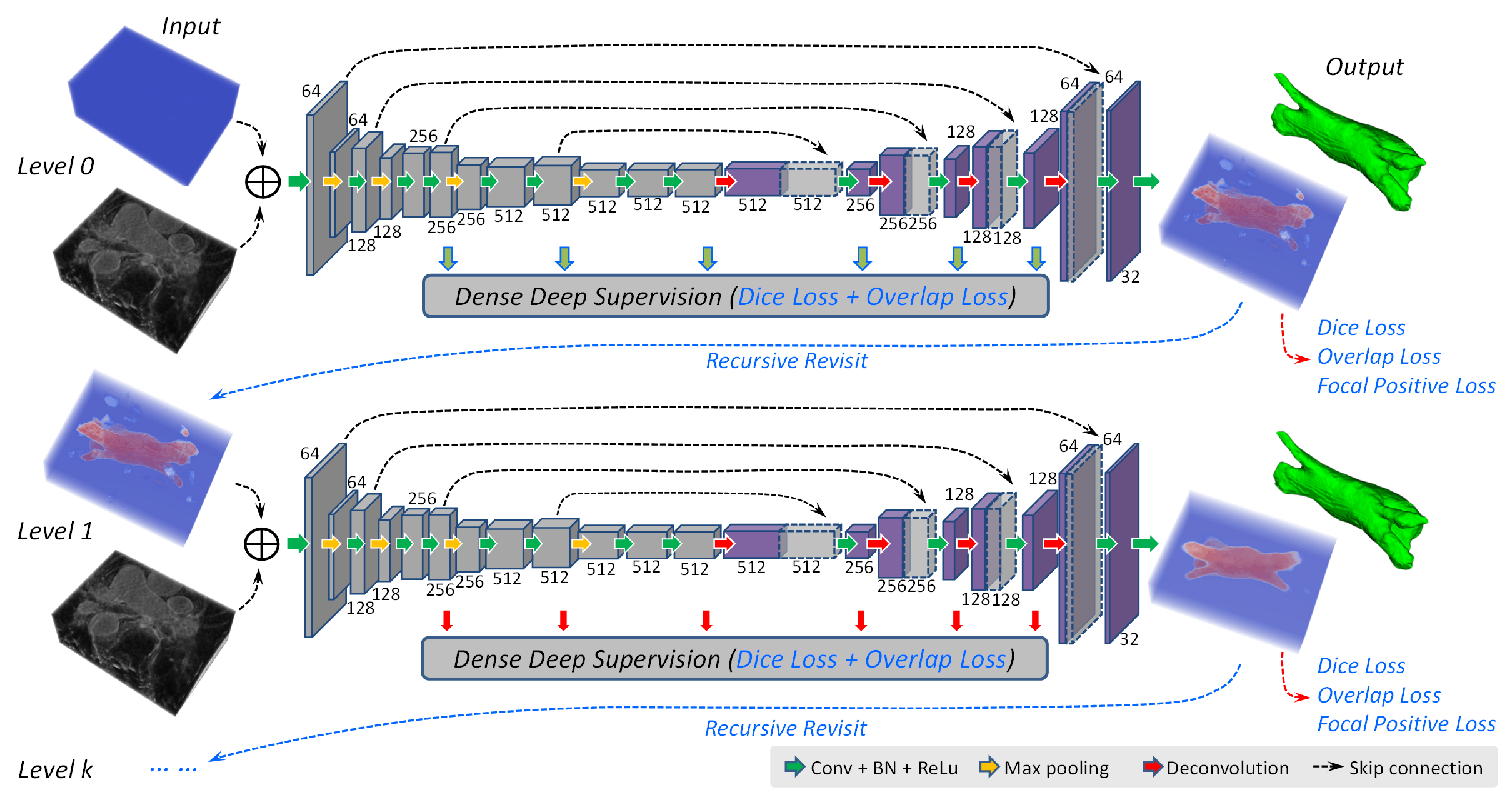}
		\caption{Schematic view of our proposed framework.}
		\label{fig:framework}
		\vspace{-0.6mm}
	\end{figure}
	
	\section{Methodology}
	Fig. \ref{fig:framework} is the schematic illustration of our proposed framework. The localized left atrium ROI serves as the input of our customized 3D network. Our network generates an intermediate segmentation and also a single channel to represent the foreground probability map. The foreground map then merges with original ROI and goes through the next refinement level. The last recursive level outputs the final segmentation of left atrium and pulmonary veins. \par

	\subsection{Localization of Left Atrium Region}
	To exclude the noise from background and narrow down the searching area, we propose to exploit a Faster-RCNN \cite{ren2015faster} network to localize the ROI of atrium and pulmonary veins. We train our Faster-RCNN with 2D slices and bounding boxes. Testing MR volume is firstly split into slices along a canonical axis and all the localized ROI in these slices are then merged into a 3D bounding box. Our ROI detection module achieves $100\%$ accuracy in hitting the atrium area. \par
		
	\subsection{Enhance the Training of 3D Network}
	Shown as Fig. \ref{fig:framework}, we customize a 3D U-net from \cite{cciccek20163d} and take it as the workhorse. To fit our proposed loss function design, we discard the last softmax layer and only output a single channel of foreground probability. We enhance the training of our network from the following three aspects. \par
	
	\subsubsection{Transfer Learning in 3D Fashion} Leveraging knowledges of well-trained model can improve the generalization ability of networks \cite{chen2015standard}. Equipped with 3D convolutions to extract spatial-temporal features, C3D model proposed in \cite{tran2015learning} is proper for the transfer learning in 3D networks. Therefore, we apply the parameters of layers \textit{conv1, conv2, conv3a, conv3b, conv4a and conv4b} in C3D model to initialize the downsampling path of our customized 3D U-net. All the layers are then fine-tuned for our atrium segmentation task. \par
	
	\subsubsection{Dense Deep Supervision (\textit{DDS})} Introducing deep supervision mechanism is effective in addressing the gradient vanishing problem faced by deep networks\cite{dou20173d}. As shown in Fig. \ref{fig:framework}, deep supervision adds auxiliary side-paths and thus exposes shallow layers to extra supervisions. Let $\mathcal{X}^{w\times h\times d}$ be the ROI input, $W$ be the parameters of main network, $w = (w^1,w^2,..,w^{S})$ be the parameters of side-paths and $w^s$ denotes the parameters of the $s^{th}$ side-path. $\tilde{\mathcal{L}}$ and $\mathcal{L}_s$ are main loss function and loss function in $s^{th}$ side-path, respectively. Components of $\tilde{\mathcal{L}}$ and $\mathcal{L}_s$ are further explained in following sections. In this work, we extend the deep supervision in a dense form, that is, we attach auxiliary side-paths in both down- and up-sampling branches and finally 6 losses in total. The final loss function $\mathcal{L}$ for our 3D U-net with dense deep supervision is elaborated in Eq. \ref{eq:ds}, where $\beta_s$ ($s\in(1,2,...,6)$) is the weight of different side-paths.
	\begin{equation} \label{eq:ds}
	\mathcal{L}(\mathcal{X};\mathcal{W},w) = \tilde{\mathcal{L}}(\mathcal{X};\mathcal{W}) + \sum_{s \in S}{\beta_s\mathcal{L}_s(\mathcal{X};\mathcal{W},w^s)} + \lambda(||\mathcal{W}||^2 + \sum_{s \in S}{||w^s||^2})
	\end{equation}
	
	\subsubsection{Dice Loss to Address Class Imbalance (\textit{DCL})} Class imbalance can bias the traditional loss function and thus make network ignore minor classes \cite{yang2017hybrid}. Dice coefficient based loss is becoming a promising choice in addressing the problem and present more clean predictions. In this work, we adopt Dice loss as a basic component for main loss and all auxiliary losses. \par
	
	\subsection{Composite Loss against Classification Uncertainty}
	\subsubsection{Overlap Loss (\textit{OVL})} Boundary ambiguity raises uncertainties for background-foreground classification. The uncertainty can be observed from the fuzzy areas in predicted probability maps. Enlarging the gap between background and foreground predictions can suppress this kind of uncertainty. In this work, we adopt the \textit{Overlap} loss (OVL) \cite{wang2018automated} to measure this kind of gap. OVL loss is a basic component of main loss and all auxiliary losses. OVL loss is defined as follows, 
	\begin{equation}
	\mathcal{L}_{ovl}(\mathcal{W},w^{s})=\sum_{i=1}^{\left| \mathcal{X}\right|}(P(y_i=1|\mathcal{X};\mathcal{W},w^{s}) * P(y_i=0|\mathcal{X};\mathcal{W},w^{s})),
	\end{equation}
	where $P$ is the predicted probability maps for foreground and background, $*$ is basic multiplication. By minimizing the \textit{Overlap} loss, our network is pushed to learn more discriminative features to distinguish background and foreground regions and thus gain confidence in recognizing ambiguous boundary locations. \par
	
	\subsubsection{Focal Positive Loss (\textit{FPL})} OVL loss focuses on enlarging the gap between foreground and background predictions. Whereas, accurately extracting foreground object, i.e. atrium and pulmonary veins, is our final task. Thus, emphasizing the foreground should be further considered. To this regard, we add a threshold map (TM) layer after probability map at the end of the main network to adaptively regularize the foreground probability map. Our network can learn to tune the TM layer and obtain voxel-specific thresholds. Finally, the TM layer can suppress weak predictions and only preserve strong positive predictions. To train the TM layer, we introduce a novel loss function, i.e. \textit{Focal Positive} loss, which is derived in a differentiable form as follows:
	\begin{equation}
	\label{eq:threshold1}
	\mathcal{L}_{fpl}(\mathcal{W},w^{s})=1-\frac{2* \left| Mask(y_i=1|\mathcal{X};\mathcal{W},w^{(t)}) * Y \right|}{\left| Mask(y_i;\mathcal{W},w^{s})\right| +\left|Y \right|},
	\end{equation}
	
	\begin{equation}
	\label{eq:threshold2}
	Mask(y_i;\mathcal{W},w^{s})=1/(1+e^{-tmp}),
	\end{equation}
	
	\begin{equation}
	\label{eq:threshold3}
	tmp=\left\lbrace \begin{array} {rcl}
	&P(y_i=1|\mathcal{X};\mathcal{W},w^{s}),&{P(y_i)>Threshold\_Map(y_i)} \\
	&-\infty,&else
	\end{array} \right. .
	\end{equation}
	By minimizing $\mathcal{L}_{fpl}$, our network can learn to enforce strong positive predictions in foreground areas defined by ground truth. $\mathcal{L}_{fpl}$ is only attached in main network. In summary, our network is trained with DCL and OVL losses existing in main network and auxiliary paths, and also the FPL loss in main network. \par

	\subsection{Recursive Refinement Scheme (\textit{RRS})}
	Probability map contains more explicit context information for segmentation than the raw MR volume. Revisiting the probability map to explore context cues for refinement is a classical strategy, like Auto-Context \cite{tu2010auto}. The core idea of Auto-Context is to stack a series of models in a way that, the model at level $k$ not only utilizes the appearance features in intensity image, but also the contextual features extracted from the prediction map generated by the model at level $k-1$. The general recursive process of an Auto-Context scheme is $\hat{y}^{k} = \mathcal{F}^k(\mathcal{J}(x, \hat{y}^{k-1}))$, where $\mathcal{F}^{k}$ is the model at level $k$, $x$ and $\hat{y}^{k}$ are the intensity image and prediction map from level $k-1$, respectively. $\mathcal{J}$ is a join operator to combine information from $x$ and $\hat{y}^{k}$. Generally, $\mathcal{J}$ is a concatenation operator. In this work, we modify $\mathcal{J}$ as a element-wise summation operator. Summation $\mathcal{J}$ enables us to reuse all the parameters of level $k-1$ in level $k$ for fine-tune. In addition, summarizing intensity image with prediction map is intuitive to highlight the target anatomical structures and also suppress the irrelevant background noise. \par

	\begin{table}[!htb] \caption {Quantitative evaluation of our proposed method} \label{table:quanti_metric}
		\centering
		\begin{tabular}{c|c|c|c|c|c}
			\toprule[2pt]
			\multirow{2}{*}{\bf{Method}} & \multicolumn{5}{c}{\bf{Metrics}}\\
			\cline{2-6}
									&Dice[\%] 		&Conform[\%]  		&Jaccard[\%]	&Adb[mm] 	&Hdb[mm] 	 \\
			\hline
			CEL						&82.015			&54.921				&69.893			&6.613		&47.396		\\
			DCL 					&84.907			&57.811				&75.291			&2.287		&30.344		\\
			DCL-DDS					&87.020			&68.033				&77.785			&2.082		&25.243		\\
			DCL-DDS-OVL				&89.833			&76.942				&81.745			&2.116		&24.591		\\
			DCL-DDS-OVL-FPL			&91.548			&81.449				&84.464			&\textbf{1.455}		&19.554		\\			
			DCL-DDS-OVL-FPL-RRS1	&92.243			&83.113				&85.639			&1.541		&18.888		\\			
			DCL-DDS-OVL-FPL-RRS2 	&\textbf{92.244}	&\textbf{83.120}	&\textbf{85.644}	&1.490	&\textbf{18.293}\\
			\hline
			\toprule[2pt]
		\end{tabular}
		\vspace{-0.2cm}
	\end{table}		
	
	\section{Experimental Results} \label{section:experiment}
	\subsubsection{Experiment materials:} We evaluated our method on the Atrial Segmentation Challenge 2018 dataset. We split the whole dataset (100 samples with ground truth annotation) into training set (80 volumes) and testing set (20 volumes). All the training and testing samples are normalized as zero mean and unit variance before inputting into network. We augmented the training dataset with random flipping, rotation and 3D elastic deform. \par	
	
	\subsubsection{Implementation details:} We implemented our framework in \textit{Tensorflow}, using 4 NVIDIA GeForce GTX TITAN Xp GPUs. We update the parameters of network with an Adam optimizer (batch size=1, initial learning rate is 0.001). Randomly cropped $144\times 64\times 144$ sub-volumes serve as input to our network. To avoid shallow layers being over-tuned during fine-tuning, we set smaller initial learning rate for \textit{conv1, conv2, conv3a, conv3b, conv4a and conv4b} as \textit{1e-6, 1e-6, 1e-5, 1e-5, 1e-4, 1e-4} . We adopt sliding window with proper overlap ratio and overlap-tiling stitching strategies to generate predictions for the whole volume, and remove small isolated connected components in final segmentation result.

	\begin{figure}
		\centering
		\includegraphics[width=1.0\linewidth]{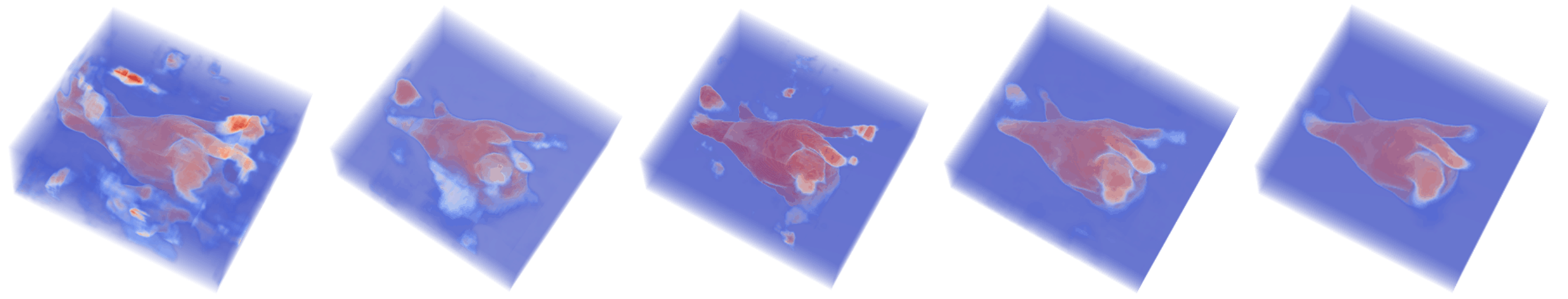}
		\caption{Visualization of probability maps generated by networks with different losses. From left to right, CEL, DCL-DDS, DCL-DDS-OVL, DCL-DDS-OVL-FPL and DCL-DDS-OVL-FPL-RRS1.}
		\label{fig:map_compare}
		\vspace{-0.5cm}
	\end{figure}

	\begin{figure}[h]
		\centering
		\includegraphics[width=1.0\linewidth]{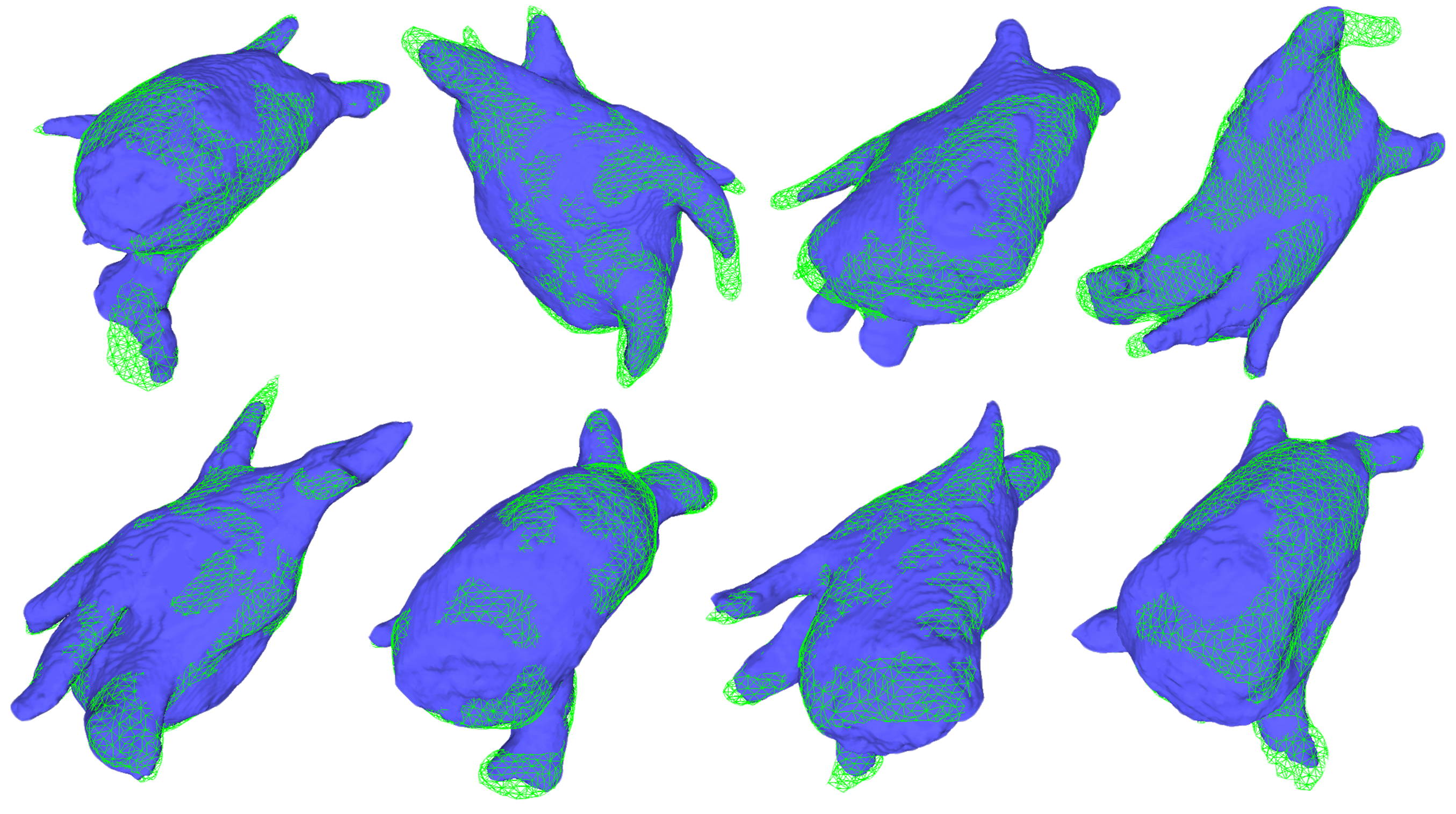}
		\caption{Visualization of our segmentation results in testing datasets. Green mesh denotes ground truth, while blue surface denotes our segmentation result. Our segmentation presents high overlap ratio with the ground truth.}
		\label{fig:seg_results}
		\vspace{-0.3cm}
	\end{figure}
	
	\subsubsection{Quantitative and Qualitative Analysis:}
	We use 5 metrics to evaluate the proposed framework, including Dice, Conform, Jaccard, Average Distance of Boundaries (Adb) and Hausdorff Distance of Boundaries (Hdb). We take the customized 3D U-net with basic cross entropy loss (CEL) as a baseline. We conduct intensive ablation study on our introduced modules, including DCL, DDS, OVL, FPL and RRS. All of the compared methods share the same basic 3D U-net architecture. Table \ref{table:quanti_metric} illustrate the detailed quantitative comparisons among different module combinations. For simplicity, all compared methods are denoted with the main module names. Both the DCL amd DDS brings improvement over the traditional CEL. Significant improvements firstly occurs as we inject the \textit{Overlap} loss. FPL further boosts the segmentation about 2 percent in Dice. Improvements brought by OVL and FPL are also verified as we visualize the foreground probability maps in Fig. \ref{fig:map_compare}. The foreground probability map becomes more sharp and noise-free as OVL and FPL involves. Mining context cues in probability maps with RRS contributes about 1 percent in Dice as two RRS levels (RRS2) are utilized. We only adopt two RRS levels to balance the performance gain and computation burden. Finally, DCL-DDS-OVL-FPL-RRS2 achieves the best performance in almost all metrics. We also visualize the segmentation results with ground truth in Fig. \ref{fig:seg_results}. Our method conquers complex variance of left atrium and pulmonary veins and achieves promising performance. \par
	
	\section{Conclusion}
	In this paper, we present a fully automatic framework for left atrium segmentation and GE-MR volumes. Originating with a network in 3D fashion to better tackle complex variances in shape and size of left atrium, we present our main contributions in introducing and verifying the composite loss which is effective in combating the boundary ambiguity and hard examples. We also propose a modified recursive scheme for successive refinement. As extensively validated on a large dataset, our proposed framework and modules prove to be promising for the left atrium segmentation in GE-MR volumes. \par
	
	\section{Acknowledgments:}
	The work in this paper was supported by a grant from the Research Grants Council of the Hong Kong Special Administrative Region (Project no. GRF 14225616), a grant from National Natural Science Foundation of China under Grant 61571304, Grant 81771598, and Shenzhen Peacock Plan under Grant KQTD2016053112051497. \par
	
	%
	\bibliographystyle{splncs}
	\bibliography{refs}

\begin{thebibliography}{10}
\providecommand{\url}[1]{\texttt{#1}}
\providecommand{\urlprefix}{URL }

\bibitem{bai2015multi}
Bai, W., Shi, W., Ledig, C., Rueckert, D.: Multi-atlas segmentation with
  augmented features for cardiac mr images. Medical image analysis  19(1),
  98--109 (2015)

\bibitem{chen2015standard}
Chen, H., Ni, D., Qin, J., et~al: Standard plane localization in fetal
  ultrasound via domain transferred deep neural networks. IEEE JBHI  19(5),
  1627--1636 (2015)

\bibitem{chen2018multiview}
Chen, J., Yang, G., Gao, Z., et~al.: Multiview two-task recursive attention
  model for left atrium and atrial scars segmentation. arXiv preprint
  arXiv:1806.04597  (2018)

\bibitem{cciccek20163d}
{\c{C}}i{\c{c}}ek, {\"O}., Abdulkadir, A., et~al: 3d u-net: Learning dense
  volumetric segmentation from sparse annotation. arXiv preprint
  arXiv:1606.06650  (2016)

\bibitem{dou20173d}
Dou, Q., Yu, L., et~al: 3d deeply supervised network for automated segmentation
  of volumetric medical images. Medical Image Analysis  (2017)

\bibitem{mcgann2013atrial}
McGann, C., Akoum, N., et~al.: Atrial fibrillation ablation outcome is
  predicted by left atrial remodeling on mri. Circulation: Arrhythmia and
  Electrophysiology pp. CIRCEP--113 (2013)

\bibitem{mortazi2017cardiacnet}
Mortazi, A., Karim, R., et~al.: Cardiacnet: Segmentation of left atrium and
  proximal pulmonary veins from mri using multi-view cnn. In: MICCAI. pp.
  377--385. Springer (2017)

\bibitem{peng2016review}
Peng, P., Lekadir, K., et~al.: A review of heart chamber segmentation for
  structural and functional analysis using cardiac magnetic resonance imaging.
  Magnetic Resonance Materials in Physics, Biology and Medicine  29(2),
  155--195 (2016)

\bibitem{ren2015faster}
Ren, S., He, K., Girshick, R., Sun, J.: Faster r-cnn: Towards real-time object
  detection with region proposal networks. In: NIPS. pp. 91--99 (2015)

\bibitem{tobon2015benchmark}
Tobon-Gomez, C., Geers, A.J., et~al.: Benchmark for algorithms segmenting the
  left atrium from 3d ct and mri datasets. IEEE TMI  34(7),  1460--1473 (2015)

\bibitem{tran2015learning}
Tran, D., Bourdev, L., Fergus, R., Torresani, L., Paluri, M.: Learning
  spatiotemporal features with 3d convolutional networks. In: ICCV. pp.
  4489--4497 (2015)

\bibitem{tu2010auto}
Tu, Z., Bai, X.: Auto-context and its application to high-level vision tasks
  and 3d brain image segmentation. IEEE TPAMI  32(10),  1744--1757 (2010)

\bibitem{wang2018automated}
Wang, Z., et~al.: Automated detection of clinically significant prostate cancer
  in mp-mri images based on an end-to-end deep neural network. IEEE TMI  (2018)

\bibitem{yang2017fully}
Yang, G., Zhuang, X., Khan, H., et~al.: A fully automatic deep learning method
  for atrial scarring segmentation from late gadolinium-enhanced mri images.
  In: ISBI 2017. pp. 844--848. IEEE (2017)

\bibitem{yang2017hybrid}
Yang, X., Bian, C., Yu, L., Ni, D., Heng, P.A.: Hybrid loss guided
  convolutional networks for whole heart parsing. In: STACOM. pp. 215--223.
  Springer (2017)

\bibitem{zheng2014multi}
Zheng, Y., et~al.: Multi-part modeling and segmentation of left atrium in c-arm
  ct for image-guided ablation of atrial fibrillation. IEEE TMI  33(2),
  318--331 (2014)

\bibitem{zhuang2010registration}
Zhuang, X., et~al: A registration-based propagation framework for automatic
  whole heart segmentation of cardiac mri. IEEE TMI  29(9),  1612--1625 (2010)

\end{thebibliography}
	
\end{document}